\documentclass[10pt,letterpaper]{article}

\usepackage{cogsci}

\cogscifinalcopy % Uncomment this line for the final submission 

\usepackage{pslatex}
\usepackage{apacite}
\usepackage{float} % Roger Levy added this and changed figure/table
\usepackage{times}
\usepackage{latexsym}
\usepackage{url}
\usepackage{kotex}
\usepackage{graphicx}
\usepackage{makecell}%To keep spacing of text in tables
\setcellgapes{3.5pt}%parameter for the spacing
\usepackage{marvosym}
\usepackage{multirow}
\usepackage{url}

\title{Text Matters but Speech Influences: \\A Computational Analysis of Syntactic Ambiguity Resolution}
 
\author{{\large \bf Won Ik Cho\textsuperscript{1} (wicho@hi.snu.ac.kr)} \\
	{\large \bf Jeonghwa Cho\textsuperscript{2} (jeonghwa@umich.edu)} \\ 
	{\large \bf Woo Hyun Kang\textsuperscript{1} (whkang@hi.snu.ac.kr)} \\
	{\large \bf Nam Soo Kim\textsuperscript{1} (nkim@snu.ac.kr)}
	\AND
  Department of Electrical and Computer Engineering and INMC, Seoul National University\textsuperscript{1}\\
  Department of Linguistics, University of Michigan, Ann Arbor\textsuperscript{2}}

\begin{document}

\maketitle

\begin{abstract}
Analyzing how human beings resolve syntactic ambiguity has long been an issue of interest in the field of linguistics. It is, at the same time, one of the most challenging issues for spoken language understanding (SLU) systems as well. As syntactic ambiguity is intertwined with issues regarding prosody and semantics, the computational approach toward speech intention identification is expected to benefit from the observations of the human language processing mechanism. In this regard, we address the task with attentive recurrent neural networks that exploit acoustic and textual features simultaneously and reveal how the modalities interact with each other to derive sentence meaning. Utilizing a speech corpus recorded on Korean scripts of syntactically ambiguous utterances, we revealed that co-attention frameworks, namely multi-hop attention and cross-attention, show significantly superior performance in disambiguating speech intention. With further analysis, we demonstrate that the computational models reflect the internal relationship between auditory and linguistic processes.

\textbf{Keywords:} 
syntactic ambiguity resolution; speech intention disambiguation; audio-text co-attention framework; prosody-syntax-semantics interface
\end{abstract}

\section{Introduction}

Resolving syntactic ambiguity is a core task in spoken language analysis, since identifying the sentence type and understanding the intention of a text form utterance is challenging for some prosody-sensitive cases. 
Notably, in some \textit{wh-in-situ} languages like Korean and Japanese, some uttered word sequences incorporate syntactic ambiguity, which leads to difficulties discerning directive speech from constative or rhetorical ones.
For example, the following sentence in Seoul Korean can be interpreted differently depending on the prosody \cite{cho2019prosody}:\medskip\\
(s1) \textbf{몇 개 가져가\hfill myech kay kacye-ka}\\
\phantom{(2) }\hfill how quantity bring-USE\footnote{Denotes an underspecified sentence ender.}\\
\phantom{(2) }(a) \textit{How many shall I take?} \\\phantom{}\hfill(LHLLH\%; \textbf{\textit{wh-}Q})\\
\phantom{(2) }(b) \textit{Shall I take some?} \\\phantom{}\hfill(LMLLH\%; \textbf{\textit{yes/no} Q})\\
\phantom{(2) }(c) \textit{Take some.} \\\phantom{}\hfill(LMLML\%; \textbf{command})\medskip\\
where L, M, and H denote relative pitch of each syllabic block and USE denotes an underspecified sentence ender. 
Unlike English translations, if given only the text with periods or question marks removed (usually provided as an output of automatic speech recognition (ASR)), the language understanding modules may not be able to determine if it is a statement or a question. Even with such marks, it is vague whether the question is \textit{yes/no} or \textit{wh-}. 

\begin{figure}
	\centering
	\includegraphics[width=0.9\columnwidth]{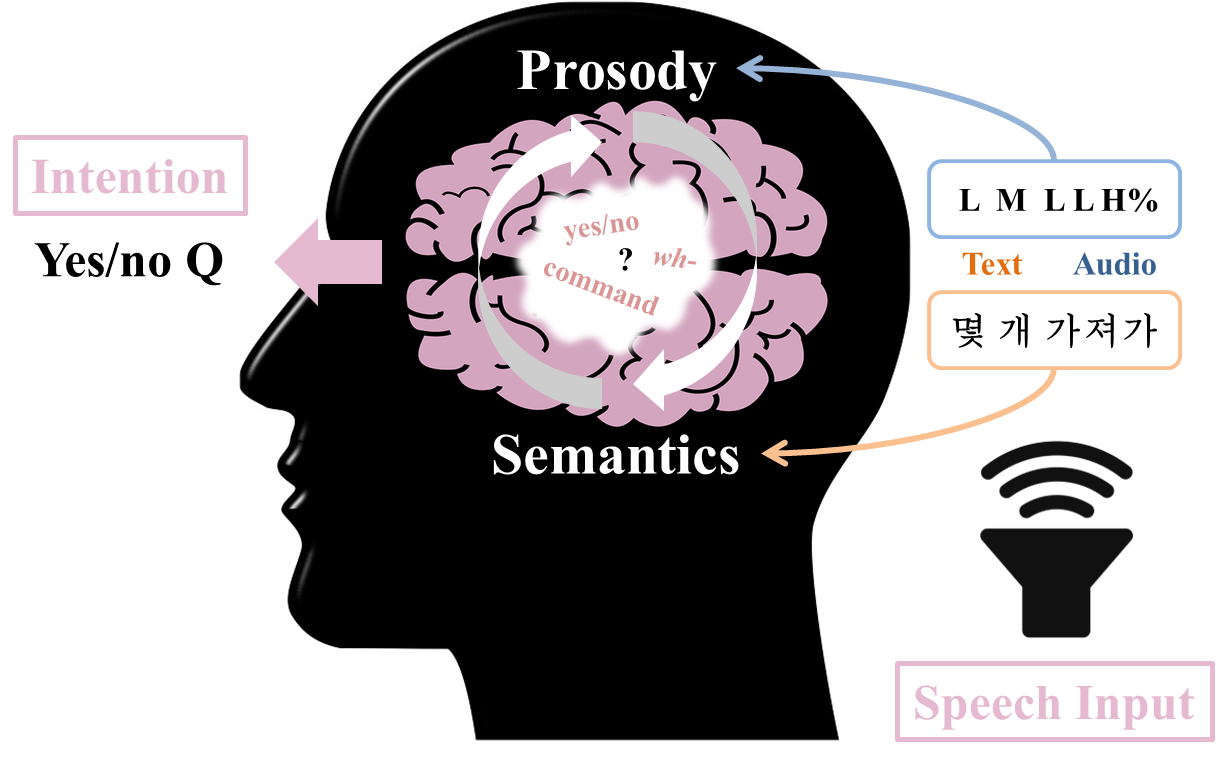}
	\caption{Prosody-syntax-semantics interface is assumed to be highly related to disambiguating confusing utterances (s1).} \label{fig:fig2}
\end{figure} 

As such, for an utterance that contains components whose roles are decided by prosody, it requires both the acoustic and textual data for spoken language understanding (SLU) modules (and even humans) to correctly infer the speech intention. In this process, the pitch sequence, the duration between words, and the overall tone all together decide the intention of an utterance. Thus, we concluded that introducing prosodic information is indispensable for resolving syntactic ambiguity, as depicted in Figure 1. 

%In this regard, we paid attention to some recent studies utilizing the Korean language, which is head-final and \textit{wh-in-situ}. 
%In L2 acquisition, \cite{kweon2004syntactic} investigated how L1 and L2 learners of Korean react differently to the syntactic ambiguity regarding double genitive terms, and \cite{baek2018prosodic} sought for the procedure in more a theoretical point of view. 
%Considering language perception, \cite{hwang2009constituent} sought for a prosody-based resolution for the dative NP ambiguity which occurs in long sentences of Korean. 
%In experimental linguistics, \cite{park2019semantic} utilized some semantic cues on resolving the ambiguity in comparative sentences of Korean, which was scrutinized theoretically in \cite{kim2010phrasal}. 
%Within speech act domain, \cite{cho2019prosody} proposed the corpus which incorporates only the syntactically ambiguous sentences regarding wh-intervention and directiveness, while the computational disambiguation has not been performed yet. We found this to be investigated more thoroughly in both engineering and scientific viewpoint.

In this paper, we first review literature on prosody-syntax interface, syntactic ambiguity resolution, and machine learning-based speech analysis. Then, we introduce the architecture of speech intention classification systems that co-utilize the audio and text features, along with a comparison with speech-based and text-aided (self-attentive) recurrent neural network (RNN) models. Next, in the experiment
section, the utilized corpus and the results are described, based on which we suggest the possibility of a connection between computational speech intention identification and human language processing mechanisms. Our contribution is as follows:

\begin{itemize}
	\item We applied parallel bidirectional recurrent encoder (P-BRE), multi-hop attention (MHA), and cross-attention (CA) to identify the speech intention of syntactically ambiguous utterances and scrutinized them in cross-modal perspective.
	\item We analyzed the result in view of experimental linguistics, linking the co-attention frameworks to neurological findings.
\end{itemize}

\section{Related Work}

%The literature on intonation dependency of speech act, syntactic ambiguity, and speech intention analysis is primarily discussed. The first one concerns phonetic property of the sentences with underspecified particles, while the second concentrates on the syntax-semantic analysis. The last one mainly considers the algorithm that deals with speech analysis, based on multi-modality.

The interaction of prosody and syntax has been long investigated regarding sentence types, especially for the question types including \textit{wh-}intervention \cite{pires2007syntax} and declarative forms \cite{gunlogson2002declarative}. Moreover, for some head-final languages, the sentence-final intonation can play a significant role in clarifying the sentences.
For instance, the prosody assigned to the final particle or word of non-scrambled Korean sentences usually decides the sentence form, e.g., declaratives or interrogatives \cite{yun2019meaning} . 
%Though the syntactic property has little correspondence with the intention of the speech, direct speech acts are identified more easily. Also, in non-canonical cases, the candidate of speech act can be restricted by the syntax \cite{cho2018speech}.

In a broader perspective, syntactic ambiguity has been dealt with not only in studies on sentences but also phrases. In Korean, dative \cite{hwang2009constituent} and comparatives \cite{kim2010phrasal} terms have been mainly investigated. 
%The diversity of interpretation incurs vagueness in inferring the syntactic structure, which comes from the difficulty of chunking and constituency parsing.
Also, linking syntax with phonetics, \citeA{baek2018prosodic} demonstrated that syntactic ambiguity is resolved via prosodic information that can elaborate the low/highness of the attachment. They handled several cases in Korean where the syntax differentiates upon phonetic properties, especially among long phrases (e.g., \textit{diligent boy's sister}). This approach is mainly concerned with contiguity theory \cite{richards2016contiguity}, which claims that syntax can make reference to phonological structure, and that movement operations can be triggered by the need to produce phonologically acceptable objects. 

%how the prosodic prominence of the spoken language differentiates the syntactic structure. In the  part,

Aforementioned studies on prosody/phonetics-syntax-semantics interface deal with various types of disambiguation, which incorporates the variance in topic, agent, experiencer, and object of syntactically ambiguous sentences. Among them, a few concerns questions, commands, and their directivity (see \citeA{yun2019meaning}). \citeA{cho2019prosody} suggested the seven-class categorization (namely statement, \textit{yes/no} question, \textit{wh-}question, rhetorical question, command, request, and rhetorical command), based on (i) sentence-middle intonation that affects topic and \textit{wh-}intervention, (ii) sentence-final intonation that changes the sentence form, and (iii) the overall tone that influences rhetoricalness. The limitation is that, the analysis beyond the categorization has yet been performed. 
For now, the prosodic activeness-based disambiguation \cite{richards2016contiguity,baek2018prosodic} well formulates the phonetic segments that clarify syntax. But we deemed it necessary to resolve the ambiguity within wider range of sentence types, promoting possibly an automatic management. In this regard, we take into account computational approaches that autonomously discover the latent and non-codified criteria.

\begin{figure*}
	\centering
	\includegraphics[width=0.8\textwidth]{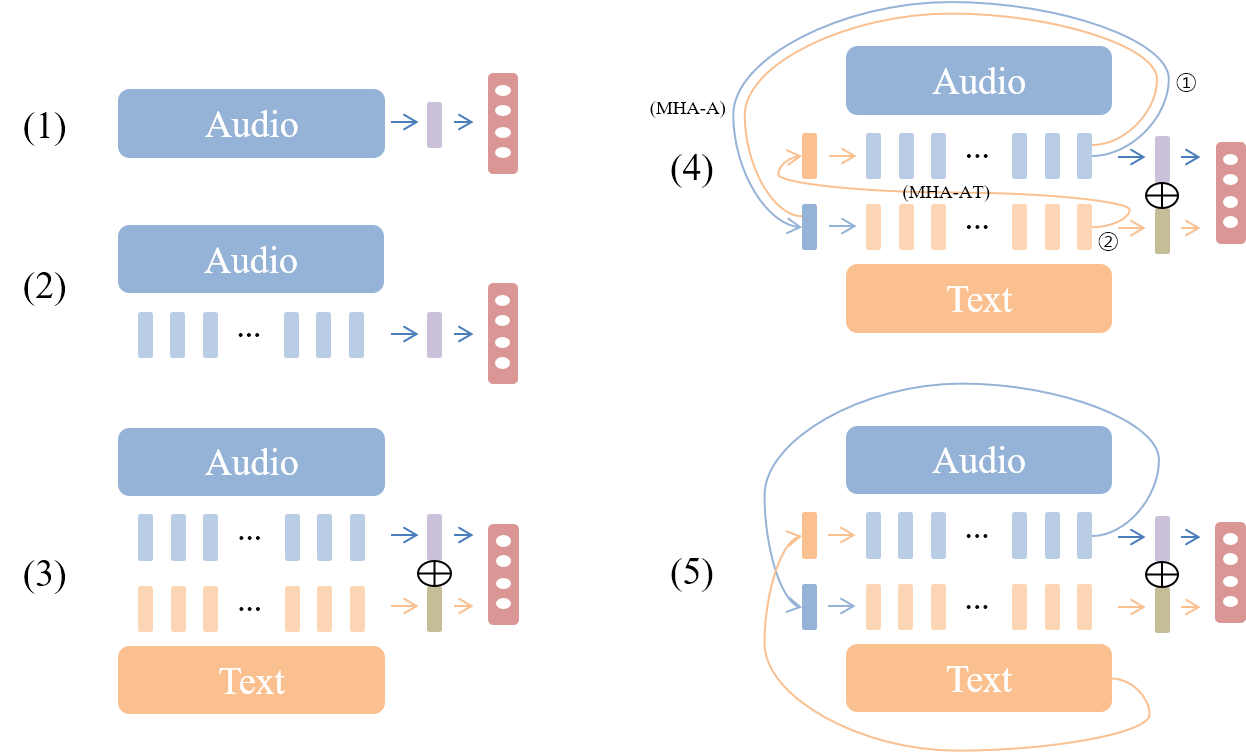}
	\caption{Block diagrams of the implemented models. Each number denotes the corresponding subsections in \textbf{Model}. Best viewed in color.} \label{fig:fig2}
\end{figure*} 

Early studies on spoken language analysis adopt a simple concatenation of acoustic and textual features \cite{gu2017speech}, where parallel convolutional or recurrent neural networks (CNN/RNNs) were used to summarize each feature. A recent study includes hierarchical attention networks (HAN) \cite{gu2018multimodal} that point out the components which are essential for inferring the answer. In the related area of speech emotion recognition, multi-hop attention (MHA) \cite{yoon2019speech} was introduced to encourage a comprehensive information exchange between textual and acoustic features. 
Nonetheless, since the experiments in literature generally utilize speech utterances with less confusing intention or emotion (e.g., syntactically non-ambiguous sentences or emotion utterances without semantic cue), there has been little study concentrating on the resolution of ambiguous sentences as in (s1). 

In terms of the prosody-syntax-semantics interface, we concluded that the interaction between acoustic and textual information is required for such cases. We aim to materialize this approach in our co-attentional architectures in the form of MHA \cite{yoon2019speech} and cross-attention (CA) \cite{lee2018stacked}, which have shown their power in the area of speech emotion recognition and image-text matching respectively.

%\begin{figure}
%	\centering
%	\includegraphics[width=0.9\columnwidth]{fig1}
%	\caption{Block diagrams of the baselines and the proposed model. Each number denotes the corresponding subsection numbers in Section 2. Best viewed in color.} \label{fig:fig2}
%\end{figure} 

\section{Model}

Here, we describe how the co-attention frameworks are constructed in terms of speech processing, self-attentive embedding, text-aided analysis, multi-hop attention, and cross-attention, as shown visually in Figure 2. In all models, the input is either audio-only (2.1-2) or audio-text pair (2.3-5). The text-only model is not taken into account since the text alone does not help resolve the syntactic ambiguity.

\subsection{Audio-only Model (Audio-BRE)}

The baseline model utilizes only audio input. Frame-level audio features are fed as an input to bidirectional long short-term memory (BiLSTM) \cite{schuster1997bidirectional}, for which the expression BRE (bidirectional recurrent encoder) is assigned \cite{yoon2019speech}. The final hidden state is fully connected to a multi-layer perceptron (MLP) to yield a correct answer as a maximum probability output in the final softmax layer. Refer to (1) in Figure 2 for an illustration of the model's architecture.
%Here, for an input frame sequence $X(t)$, RNN and MLP is defined as following:
%The specification of the model architecture is to be provided afterwards.

\subsection{Audio-only Model with Self-attentive Embedding (Audio-BRE-Att)}

Since the audio-only BiLSTM\footnote{In this paper, we interchangeably use BRE and BiLSTM.} model lacks information regarding the identification of the core parts in analyzing an utterance, we augmented a self-attentive embedding layer as utilized in the sentence representation \cite{lin2017structured}. In brief, a context vector, which has the same width as the hidden layers of the BiLSTM, is jointly trained to assign  weight vector to the hidden layer sequence thereof. 
%To be specific, the weight (attention) vector is obtained by a  column-wise multiplication of the context vector and the hidden layers. All the hidden layers are summed to make up a final feature layer, which is fed as an input of MLP.

The whole process, as in (2) of Figure 2, implies that the weight is decided upon the overall distribution of acoustic features. Since the acoustic feature reflects the lexicon and the syntactic property, the weight ends up playing a crucial role in predicting the intention of the speech.  
%The detailed scheme is described as:
%Refer to (2) in Figure 1 for illustration.

\subsection{Parallel Utilization of Audio and Text (Para-BRE-Att)}

Unlike emotion analysis where either textual or acoustic features do not necessarily dominate, in intention analysis, obtaining textual information can bring a significant advantage \cite{gu2017speech}, even when a period or a question mark is omitted as in our experiment. Here, text input for the ambiguous sentences are identical (without punctuation mark) for two to four different versions of the speech, but feeding them as an input of separately constructed \textit{Audio-BRE} may provide supplementary information. 
The final hidden layer of \textit{Audio-BRE-Att} is concatenated with that of the \textit{Text-BRE-Att} (BRE-Att that exploits textual features) to make up a new feature layer, as illustrated in (3) of Figure 2. 

%\subsection{Fusion of text-aided approach and self-attentive embedding}
%
%To boost the utility of attention mechanism in jointly training, we simultaneously applied self-attentive embedding layer to both speech and text RNN. The sum of hidden layers of both self-attentive RNN were concatenated to make up a new final feature layer. This is a plausible extension of models constructed in Section 2.2 and 2.3, and may successfully project textual information in the inferring process that mainly utilizes the acoustic features. However, there is a significant disadvantage in this model that the prosodic information is not fully utilized in providing attention to each token in the text, and vice versa.

\subsection{Multi-Hop Attention (MHA)}

In multi-hop attention (MHA) \cite{yoon2019speech}, which is proposed for speech emotion recognition, textual and acoustic features interact by sequentially transmitting information to each other. This is the background of the expression `multi-hop', where the hopping is performed by adopting the final representation of each feature as a context vector of the other as in (4) of Figure 2. The final output of the former and the latter are eventually concatenated. 

Here, we first implement hopping only \textit{from audio to text} (4a) (MHA-A), and then augment \textit{from text to audio} to make up (4b) (MHA-AT). They showed better performance than the further hopped model (i.e., MHA-ATA) in the original study \cite{yoon2019speech}. Also, it is empirically more acceptable than the reverse case (e.g., MHA-T/TA) since  auditory sensory first faces acoustic data than semantic information.%\footnote{Here, `auditory sensory'  represents the blue bars in Fig. 2, and the lines are the information that is transmitted to determining the attention to the hidden layers of other modality.}.  

\subsection{Cross-Attention (CA)}

From the perspective of another co-attention framework, we adopt cross-attention (CA) that fully utilizes the information flow exchanged simultaneously by both acoustic and textual features, as depicted in (5) of Figure 2. 
In the preceding paper on image-text matching \cite{lee2018stacked}, image segments are utilized in determining the attention vector for the text, and similarly in reverse.
%Here, cross-attention implies that two (or possibly more) sequential features are aligned to provide useful information to each. 
Thus, not limited to using the representation regarding one feature as a context vector of the other's attention weight,  we assumed it also plausible to utilize the final representation of \textit{Audio-BRE-Att} in making up a weight vector for \textit{Text-BRE} and vice versa. In this case, self-attentive embedding was not applied to the textual features, in order to reflect the auditory-first nature\footnote{Since auditory sensory meets the speech before the audio is encoded to lexical components, we considered it fair to assign different levels of representation and weight regarding both modalities. Here, we implement it in the way of giving self-attentive embedding only to the audio features. In fact, providing attention to the textual features as well, resultingly degraded the performance; we also tried to avoid that case.}. %and avoid the analysis being performed on audio and text by an equal rate.

\section{Experiment}

\subsection{Corpus Description}

The dataset for the analysis of ambiguous speech, which requires prosodic disambiguation, is a corpus that contains about 1.3K sentences with two to four different types of prosody (and corresponding intention). Specifically, each sentence  (i) starts with a \textit{wh-}particle, (ii) incorporates a predicate made up of general verbs and pronouns, and (iii) ends with underspecified sentence enders so that the overall prosody varies according to intention (and sometimes with politeness suffix). All the sentences received consensus of three Korean native speakers, and the total number of speech utterances reaches 3,552. A male and a female speakers recorded each utterance with an appropriate prosody for each intention, to obtain a dataset of size 7,104. 
The number of intentions is seven, namely statement (S), yes/no question (YN), \textit{wh-}question (WH), rhetorical question (RQ), command (C), request (R), and rhetorical command (RC). The categorization is slightly modified from  the publicly available Korean corpus \cite{cho2018speech} to reflect \textit{wh-} intervention as illustrated in (s1). 
The specification of the corpus is given in Table 1. 
The corpus\footnote{\url{https://github.com/warnikchow/prosem}} and its detailed generation scheme are published as a separate article \cite{cho2019prosody}.
%The details of the corpus and the generation scheme are available in \citeA{cho2019prosody}.

\begin{table}[]
	\centering
	\makegapedcells
	\resizebox{0.8\columnwidth}{!}{%
		\begin{tabular}{|c|c|c|c|c|c|c|c|}
			\hline
			& \textit{\textbf{S}} & \textit{\textbf{YN}} & \textit{\textbf{WH}} & \textit{\textbf{RQ}} & \textit{\textbf{C}} & \textit{\textbf{R}} & \textit{\textbf{RC}} \\ \hline
			\textit{\textbf{Who}} & 547 & 544 & 446 & 202 & 112 & 26 & 18 \\ \hline
			\textit{\textbf{What}} & 294 & 283 & 186 & 64 & 32 & 14 & 4 \\ \hline
			\textit{\textbf{Where}} & 64 & 64 & 49 & 6 & 11 & 4 & 1 \\ \hline
			\textit{\textbf{When}} & 37 & 54 & 40 & 22 & 0 & 4 & 15 \\ \hline
			\textit{\textbf{How}} & 59 & 62 & 28 & 8 & 6 & 0 & 0 \\ \hline
			\textit{\textbf{How much}} & 84 & 40 & 100 & 0 & 14 & 8 & 0 \\ \hline
		\end{tabular}%
	}
	\caption{A frequency matrix on \textit{wh-} particles and the intention types, excerpted from \cite{cho2019prosody}. For male and female cases each, the total speech files were randomly organized and split into train and test set with the ratio of 9:1.}
\end{table}

\subsection{Features and Architecture}

%\subsubsection{Features} 

For acoustic features, mel spectrogram (MS) and root mean square energy per frame (RMSE) were obtained by Librosa and were concatenated frame-wisely. 
%The two sequences were augmented to make up an input matrix of size 129 (dim) * 600 (frame). 
For textual features, %character-level embeddings were utilized. For the character-level features, 
two types of character-level representation were adopted, namely sparse and dense, as they show best performance for classification tasks \cite{cho2019investigating}. For sparse vectors, multi-hot encodings of the Korean characters were used  \cite{song2018sequence}. 
These features display conciseness and also preserve the property of the blocks as a conjunct form. For dense features that regard distributional semantics, the recently disclosed fastText \cite{bojanowski2016enriching}-based word vector dictionary was exploited \cite{cho2018real}.
%, resulting in input size of $L$ (dim) * 30 (characters) where $L$ depends on the embedding methodology. 

Considering the head-finality of the Korean sentences, we embedded the acoustic and textual features backward from the endpoint\footnote{This was controversial, but was to assure that all the models can pay attention to the sentence end of each utterance, where the underspecified functional particles lie in. This does not mean \textit{uni-LSTM}; this instead means that the characters are placed from the end slot of the sequence in the input-level.}. The maximum sequence length was fixed to cover the longest input, where all the features with shorter frame/character length were padded with zeros.
%\begin{figure}
%	\centering
%	\includegraphics[width=\columnwidth]{fig2}
%	\caption{Blocks for a simple description of implementation.} \label{fig:fig2}
%\end{figure} 
%\subsubsection{Character-level embedding}
%For character-level text embedding, two types of representation were utilized; namely sparse \cite{song2018sequence} and dense \cite{cho2018real}, as shown best performing for classification tasks. For sparse vectors, multi-hot encodings of the Korean syllabic blocks were used. They display conciseness and also present preserve property of the blocks as a conjunct form character. For dense features, recently distributed fastText \cite{bojanowski2016enriching}-based word vector dictionary was exploited. %, trained with about 2M drama lines which contains various conversation-style and non-canonical utterances. We expect the latter to display the distributive semantics that may provide additional knowledge in the training phase.
%\subsubsection{Model construction and parameters}
Models were implemented with Keras \cite{chollet2015keras}, using TensorFlow backend \cite{tensorflow2015-whitepaper}. Architecture and hyper-parameter specification are provided separately on-line, with all the codes for implementation\footnote{\url{https://github.com/warnikchow/coaudiotext}}.%\footnote{\url{https://github.com/warnikchow/coaudiotext}}.

\begin{table}[]
	\centering
	\makegapedcells
	\resizebox{\columnwidth}{!}{%
		\begin{tabular}{|c|c|l|c|c|}
			\hline
			\multicolumn{1}{|l|}{\multirow{2}{*}{}} & \multicolumn{2}{c|}{\textbf{Accuracy (F1)}}                 & \multirow{2}{*}{\textbf{Param.s}} & \multirow{2}{*}{\textbf{Comp.}} \\ \cline{2-3}
			\multicolumn{1}{|l|}{}                  & \textbf{Sparse}       & \multicolumn{1}{c|}{\textbf{Dense}} &                                   &                                 \\ \hline
			\textbf{(1) Audio-BRE}                  & \multicolumn{2}{c|}{83.9 (0.652)}                           & 116K                              & 65s                             \\ \hline
			\textbf{(2) Audio-BRE-Att}              & \multicolumn{2}{c|}{89.3 (0.759)}                           & 190K                              & 67s                             \\ \hline
			\textbf{(3) Para-BRE-Att}               & \textbf{93.2 (0.919)} & 92.8 (0.919)                        & 260K                              & 70s                             \\ \hline
			\textbf{(4a) MHA-A}                     & \textbf{93.8 (0.928)} & \textbf{93.5 (0.922)}               & 266K                              & 67s                             \\ \hline
			\textbf{(4b) MHA-AT}                    & 92.8 (0.909)          & 91.8 (0.904)                        & 270K                              & 67s                             \\ \hline
			\textbf{(5) CA}                         & 91.8 (0.884)          & \textbf{93.5 (0.919)}               & 326K                              & 65s                             \\ \hline
			\textbf{(3') Para-ASR}                  & 90.0 (0.822)          & \multicolumn{1}{c|}{-}              & -                                 & -                               \\ \hline
			\textbf{(4a') MHA-ASR}                  & 90.2 (0.799)          & \multicolumn{1}{c|}{-}              & -                                 & -                               \\ \hline
		\end{tabular}%
	}
	\caption{Result on the 10\% test set. For each model, where all reached the convergence, we chose the intersection among 5-best accuracy and F1 checkpoints that were yielded during the first 100 epochs of training.}
	\label{tab:my-table}
\end{table}

\subsection{Result}

Table 2 shows the comparison result utilizing the corpus dataset. Both train and test sets in (1-5) incorporate the scripts of ground truth, and for the others, the test set scripts were ASR result. Input materials are either sole audio or audio-text combined, both in the training and test phase. \smallskip\\
\textbf{Attention matters:} First, in (1) and (2), we observed that audio itself incorporates substantial information regarding speech intention, and physical features such as duration, pitch, tone, and magnitude can help yield semantic understanding via attention mechanism. This seems to be related to the phenomenon where people often catch the underlying intention of a speech even when they fail to understand the whole words \cite{hellbernd2016prosody}. Also, it was shown that attaching the attention layer guarantees stable convergence of the learning curve in the training phase. \smallskip\\
\textbf{Text matters:} Next, as expected, the text-aided models (3-5) far outperform the audio-only ones (1, 2), notwithstanding bigger trainable parameter set size and the computation time. Although the character-level features we utilized do not necessarily represent semantic information (which is held at least at morpheme-level), this result can be interpreted as implying that utilizing textual features can help recognize the prosodic prominence within the audio features \cite{talman2019predicting}. It was beyond our expectation that the sparse vectors outperform the dense ones in general. The exception was in CA, which implies that CA takes more advantage from the distributional semantics within the text embedding. We infer that CA may exhibit significance if the utterances become more cumbersome, where pre-trained language models (LMs) prevalent these days might be helpful.\smallskip\\
%seems to originate in the distortion caused by the dense character vectors that regard polysemy.\smallskip\\
\textbf{Co-attention framework helps:} In (3-5), we noticed that co-utilizing both audio and text in making up the attention vectors as in (4) \textit{MHA} or (5) \textit{CA} shows better performance than a simple concatenation in \textit{Para-BRE-Att}. Since the studies on speech emotion analysis \cite{pell2011emotional,ben2016prosody} claim that prosody and semantic cues cooperatively affect inferring the ground truth, we suspect that a similar phenomenon takes place in the case of speech intention. That is, \textit{acoustic and textual processing significantly benefits from a consequent or simultaneous interaction with each other}. \smallskip\\%Especially, (4a) outperforming the others implies that emotion analysis-based strategy also works well with the intention analysis, again connoting that both process highly regard speech prosody of which the region of activation in brain is expected to be similar. In general, 
\textbf{Over-stack may bring a collapse:} We first hypothesized that (4b) or (5) would show better performance compared to (4a) due to a broader or deeper exchange of information between both sources. Instead, we observed performance degeneration, leading to the conclusion that the inference becomes unstable if too much information is stacked. It is assumed that speech intention analysis is \textit{affected dominantly by the combination of speech analysis and a speech-aided text analysis} (4a, 5), preferably with the smaller contribution of text-aided speech analysis (4b), though the performance of the models may not be directly linked to actual human processing mechanism. This shows that \textit{text matters but speech influences}, as will be discussed further below.\smallskip\\
%it difficult to compare between (4b) and (5), though (4b) seems more biologically plausible. It is clear that the acoustic information hits the brain first and the textual analysis follows, but it is still in question that the analysis on the acoustic feature can be done independently from the textual analysis. 
%Also, by (d), we observed that the proposed model provides more efficient structure for inducing and combining acoustic and textual information, which is more reasonable considering the psycho-linguistic nature. Moreover, (e) proves that the cross-alignment finally enhances the performance even given the interaction of multiple attention sources.
\textbf{In-depth analyses:} For a practicability of the systems, model parameter size and training time per epoch were recorded (Table 2). Taking into account that audio processing itself incorporates huge computation, co-utilizing the textual information seems to bring significant improvement. 

Then, we performed an additional experiment on ASR result\footnote{ASR was performed with a freely available API:\\ \url{https://aibril-stt-demo-korean.sk.kr.mybluemix.net/}}
(3', 4a'), especially for the test utterances, where (3) and (4a) were chosen to observe how the degeneration differs in concatenation and co-attention frameworks. The training was performed with the ground truth, and the models for the sparse textual features were chosen upon the result with it as well (3, 4a). 
It is notable that both perform competitively with the case of perfect transcription, but the degeneration was more significant in the co-attention framework. This implies that the framework utilizing textual information more aggressively is ironically more vulnerable to errors. Thus, precise ASR and error-compensating text processing are both required for the improvement and application of the system.

Lastly, we observed that (i) the co-utilization of acoustic and textual features shows strength in identifying the intention classes that are highly influenced by prosody itself, e.g., distinguishing RQs from pure questions. Some cases deeply concerned the lexicon, e.g., distinguishing commands from statements or requests from \textit{yes/no} questions. (ii) On the other hand, figuring out \textit{wh-} intervention between \textit{yes/no} and \textit{wh-} questions, depended more on the interaction of audio and text processing, shown by a superior performance of  \textit{MHA} than \textit{Para-BRE-Att}. 
These two observations can be explained as follows: in Korean spoken language, identifying rhetoricalness often accompanies non-neutral emotion (as suggested for a syntactically similar language \cite{miura1995production}), whereas  \textit{wh-}intervention mostly involves phonological properties. 
%Also, directiveness of an utterance is frequently determined by the semantic roles of predicates, which fill the dropped subject. 
Thus, we assume that (i) the emotion-related identification concerns a comprehensive understanding of the utterance as in \textit{Para-BRE-Att}, while (ii) the elaborate processing of verbal data requires an analysis that pays more attention to the details of audio and text.  %; especially given the multi-hot encoded vectors which have robustness against errors. This implies that the co-utilization models helps mitigate the effect of ASR errors with the help of audio source, finally helping better infer the correct intention. 

\section{Explainability}

We want to claim that unlike emotion recognition, which is often dominated by the voice tone, intention analysis should rely on both acoustic and textual features. More precisely, the genuine intention cannot be inferred unless the audio and text are both given if the sentence incorporates syntactic ambiguity. This is the background for this study to incorporate various co-attention frameworks. 

Neurological evidence also supports the idea of incorporating both acoustic and textual information in sentence comprehension. For example, when we try to comprehend a speech, the parts of the brain that deal with understanding the meaning (e.g., whether it is a statement, a question, or a command) include mainly the areas which take charge of syntax-semantics \cite{tanner2007redefining} and some specific regions of the temporal lobe that incur disorder in phonological decoding if damaged \cite{boatman2000transcortical}. At the outset, word forms are recognized based on the decoding process of auditory cortices \cite{dewitt2012phoneme}, and it is known that further language processing engages the Wernicke's area \cite{dewitt2013wernicke}. Thus, it is not unnatural to consider the further language processing as the pipeline, which corresponds with our observation that the attention regarding speech precedes that of lexicons.
However, we want to claim that not only are such phonological and lexical processing subsequent, but also the related regions interchange the information with each other via various language pathways \cite{friederici2012cortical}, and possibly do so more intensively when faced with ambiguous utterances. 

%latter is are excited by the stimulus that comes through auditory  
%Broca's area (related with linguistic prosody \cite{belyk2013perception}).

Moreover, an interaction between different lobes or hemisphere of the brain is also assumed, for example via dorsal and ventral pathways \cite{friederici2012cortical} or corpus callosum \cite{sammler2010prosody}. Here, the information flow is not only restricted to phonological information but can be extended to emotional information \cite{schmidt2013emotional}, as required for understanding rhetorical utterances. Supported by this biological phenomena and our experimental result, we can confirm that the correct identification of syntactically ambiguous sentences can benefit from the active co-utilization of acoustic and textual information.%, although the latter is more expected to be influenced by the former due to the physiological nature.

\section{Application}

As stated previously, ambiguous utterances are disturbing factors for speech intention understanding, which can mislead a computational model to provide a wrong intent or item. However, aggregating both audio and text actively in analyzing such utterances can help more precisely predict the intention, if given a transcription with high accuracy. We are optimistic that our approach will prove meaningful for solving intriguing problems. 
In real life, co-attention frameworks can help machines or aphasia patients understand speech. Followingly, the system users or social chatbots may be able to provide proper response/reaction in free-style or goal-oriented conversations with others.

%In a slightly different viewpoint, the proposed model can also be utilized by patients who find it difficult to understand  the emotion or intent conveyed by voice tone and prosody. The model may recognize the emotion and intention of the speaker and report it to the users so that they can make a proper reaction/response. 
Beyond the intention-related syntactic ambiguity, the implemented structures can be utilized in other kinds of natural language processing systems that incorporate multi-modal inputs that are expected to be interactive with each other. For example, the proposed network can be utilized in providing a proper translation in multi-modal context. Beyond just a text-to-text transformation, the system might capacitate abstracting and utilizing the source speech or image as an auxiliary input for the conventional machine translation process.

\section{Conclusion}

In this paper, we constructed a speech intention recognition system using co-attentional frameworks inspired by psycholinguistics and prosody-semantics interface of human language understanding. Multi-hop attention and cross-attention outperformed the conventional speech/attention-based and text-aided models, as shown by the evaluation using the audio-text pair recorded with manually created scripts. An additional experiment with ASR output was also conducted to guarantee real world usage. % and placed the room for the improvement in text processing.
The implemented systems can help SLU modules correctly infer the intention of syntactically/semantically ambiguous utterances in Seoul Korean and possibly in a multi-lingual manner. Besides, we hope the results provide empirical evidence for finding out the language processing mechanism of ambiguous utterances. %people who find it difficult to extract emotion/intention from voice to properly understand the meaning.

\section{Acknowledgments}

%To be added.

%This research was supported by Projects for Research and Development of Police science and Technology under Center for Research and Development of Police science and Technology and Korean National Police Agency funded by the Ministry of Science, ICT and Future Planning (PA-J000001-2017-101). Also, t
This work was supported by the Technology Innovation Program (10076583, Development of free-running speech recognition technologies for embedded robot system) funded By the Ministry of Trade, Industry \& Energy (MOTIE, Korea). 
%This work was supported by the research fund of Signal Intelligence Research Center
%supervised by Defense Acquisition Program Administration and Agency for Defense Development of Korea.
The authors appreciate Jeemin Kang for the great support in making up the dataset. Also, we are thankful to the anonymous reviewers for providing helpful comments.

%In the \textbf{initial submission}, please \textbf{do not include
%  acknowledgements}, to preserve anonymity.  In the \textbf{final submission},
%place acknowledgments (including funding information) in a section \textbf{at
%the end of the paper}.
%
%
%\section{References Instructions}
%
%Follow the APA Publication Manual for citation format, both within the
%text and in the reference list, with the following exceptions: (a) do
%not cite the page numbers of any book, including chapters in edited
%volumes; (b) use the same format for unpublished references as for
%published ones. Alphabetize references by the surnames of the authors,
%with single author entries preceding multiple author entries. Order
%references by the same authors by the year of publication, with the
%earliest first.
%
%Use a first level section heading, ``{\bf References}'', as shown
%below. Use a hanging indent style, with the first line of the
%reference flush against the left margin and subsequent lines indented
%by 1/8~inch. Below are example references for a conference paper, book
%chapter, journal article, dissertation, book, technical report, and
%edited volume, respectively.

\bibliographystyle{apacite}

\bibliography{my_bib_warnik}

\end{document}